\documentclass{article}
    \usepackage[preprint]{neurips_2020}

     \usepackage{neurips_2020}

\long\def\ignorethis#1{}

\newcommand{\real}{\mathbb{R}}

\usepackage[utf8]{inputenc} 
\usepackage[T1]{fontenc}    
\usepackage{hyperref}       
\usepackage{url}            
\usepackage{booktabs}       
\usepackage{amsfonts}       
\usepackage{nicefrac}       
\usepackage{microtype}      
\usepackage{graphicx}
\usepackage[ruled]{algorithm2e}
\usepackage{algorithmic, multicol}
\usepackage{graphicx}
\usepackage{lipsum}
\usepackage{amsfonts}
\usepackage{amsmath}
\usepackage{amsthm}
\usepackage{wrapfig}
\usepackage{booktabs}

\theoremstyle{definition}

\usepackage[x11names]{xcolor}

\newcommand{\defeq}{\mathrel{\overset{\makebox[0pt]{\mbox{\normalfont\tiny\sffamily def}}}{=}}}
\definecolor{myBlue}{rgb}{0,0,0.3}

\hypersetup{
	colorlinks=true,
	linkcolor=blue,
	filecolor=myBlue,      
	urlcolor=myBlue, 	
	citecolor=myBlue, 	
}

\renewcommand{\vec}[1]{\boldsymbol{#1}}
\newcommand{\rpm}{\sbox0{$1$}\sbox2{$\scriptstyle\pm$}
  \raise\dimexpr(\ht0-\ht2)/2\relax\box2 }

\def\statespace{\mathcal{S}}
\def\actionspace{\mathcal{A}}

\def\observationspace{\mathcal{O}}
\def\observation{o}

\newcommand\blfootnote[1]{%
  \begingroup
  \renewcommand\thefootnote{}\footnote{#1}%
  \addtocounter{footnote}{-1}%
  \endgroup
}

\title{Learning Causal Models Online}

\author{%
	Khurram Javed*, Martha White\\
RLAI Lab,\\
University of Alberta\\
\And
	Yoshua Bengio\\
MILA, \\ Université de Montréal\\
}

\begin{document}

\maketitle

\begin{abstract}
Predictive models -- learned from observational data not covering the complete data distribution -- can rely on spurious correlations in the data for making predictions. These correlations make the models brittle and hinder generalization. One solution for achieving strong generalization is to incorporate causal structures in the models; such structures constrain learning by ignoring correlations that contradict them. However, learning these structures is a hard problem in itself. Moreover, it's not clear how to incorporate the machinery of causality with online continual learning. In this work, we take an indirect approach to discovering causal models. Instead of searching for the true causal model directly, we propose an online algorithm that continually detects and removes spurious features. Our algorithm works on the idea that the correlation of a spurious feature with a target is not constant over-time. As a result, the weight associated with that feature is constantly changing. We show that by continually removing such features, our method converges to solutions that have strong generalization. Moreover, our method combined with random search can also discover non-spurious features from raw sensory data. Finally, our work highlights that the information present in the temporal structure of the problem -- destroyed by shuffling the data -- is essential for detecting spurious features online. 
  
\end{abstract}

\section{Introduction}
\nocite{Tange2011a}
\blfootnote{*Correspondence at kjaved@ualberta.ca. Work done while visiting MILA.}
Over the past decade, we have realized several milestones associated with artificial intelligence by minimizing empirical risk on raw data \citep{krizhevsky2012imagenet, mnih2015human}.
When sufficient data covering the complete support of the data distribution is available, minimizing empirical risk leads to a reasonable solution. However, oftentimes we want to learn from one part of the data-distribution and generalize to another. This could be due to two reasons: First, the real data distribution could be so large that it is infeasible to collect data covering the complete distribution. Second, it could be hard to access parts of the distribution -- such as collecting data for testing a parachute for landing a rover on Mars. These cases require an extreme form of generalization: systematic zero-shot generalization. It is unlikely that we would achieve such generalization by minimizing empirical risk on a small part of the data distribution.

One potential solution for achieving systematic generalization is to learn a causal structure about the world \citep{pearl2009causality, lopez2016dependence,lake2017building}. A causal structure can constrain the dependence between variables of the world, weeding out spurious correlations. 
Unfortunately, learning the causal structure in the general case requires collecting data covering the complete distribution. On the surface, incorporating a causal structure just pushes the problem of learning a predictive model that generalizes to an equally hard problem of learning variables and the correct causal structure between those variables.

\begin{figure}
  \centering
  \includegraphics[width=0.90\textwidth]{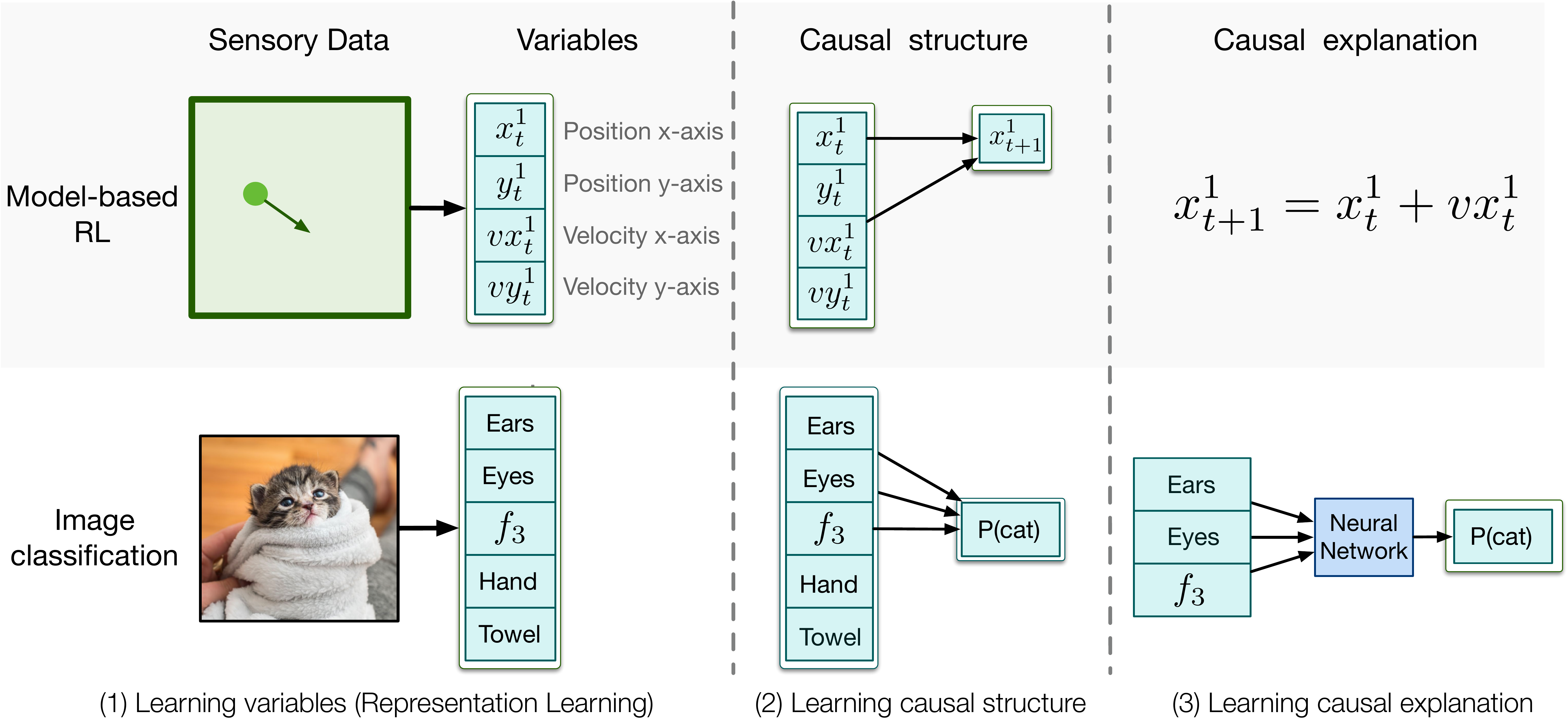}
  \caption{We look at two examples of potential causal models: predicting the position of a bouncing ball, and predicting the class of an image. For both cases, a causal model has three components: (a) extracting a set of abstract variables -- features -- from the raw sensory data; (b) removing the spurious variables from the list of variables; (c) capturing the exact relation between the causal variables and the target. The variables don't necessarily correspond to interpretable aspects of the world. They can represent uninterpretable abstractions, as $f_3$ for the cat image.}
  \label{threecomponents}
\end{figure}

To learn a causal model from observations, an agent has to learn three important components. First, it has to learn variables of the world from raw sensory data. These variables, or abstractions, can be interpretable -- balls, laptops, color -- or uninterpretable phenomena. In machine learning, learning these variables from raw data is termed \emph{representation learning}. Second, the agent has to learn the causal relations between these variables. For instance, the agent might have to learn that there is no causal relation between the presence of grass in an image and classification of an animal as a cow -- the same animal standing on a beach is still a cow. Finally, an agent has to understand the interactions between these variables for making predictions. Knowing that the position of a car in the future depends on its velocity and not color is not sufficient; the agent has to learn the position at time $t+1$ equals the sum of position and velocity at time $t$. All three of these components, when combined, constitute a potential causal model as shown in Fig.~\ref{threecomponents}. The potential causal model may or may not be the true causal model. We define a potential causal model to be the \textit{true causal model} of a target $y$ \textit{iff} it can correctly predict $y$ on the complete data-distribution.

Deep neural networks are capable of representing all three components of a causal model in their parameters. A neural network can transform raw sensory data -- images, audio -- into abstract features represented by activations. It can also model the relations between these features. For instance, by setting weight from one feature to another to zero, the neural network can encode that there is no causal dependence between the two features. Finally, a linear function -- last layer of a neural network -- can combine the features to make predictions. 

Even though deep neural networks can represent a potential causal model in their parameters, training a neural network by minimizing empirical risk on a small part of the data-distribution is unlikely to recover the true causal model. However, neural networks combined with the right learning algorithms and data might be sufficient for discovering the true causal model.

Perhaps the most effective method for discovering causal mechanisms about the world from observational data is the scientific method \citep{sep-scientific-method}. It has allowed us to discover simple causal relations -- acceleration due to gravity is independent of the mass of the body -- to more complex ones -- micro-organisms invisible to the naked eye cause infections. The knowledge discovered through the scientific method also has strong generalization; we can use this knowledge to build rovers that can land and operate on a planet hundreds of millions of miles away.  Instead of searching for the correct causal model directly, the scientific method works by continually testing hypotheses and rejecting those that contradict the data. Models that are not falsified are treated as likely true. 
 We take a similar approach -- instead of finding causal variables and structures directly, we design a scalable online learning algorithm that continually detects and removes spurious features from a neural network model with the hope that the leftover model is likely the causal one.


\begin{wrapfigure}{R}{0.55\textwidth}

	\vspace{-20pt}
	\begin{center}
		\includegraphics[width=0.50\textwidth]{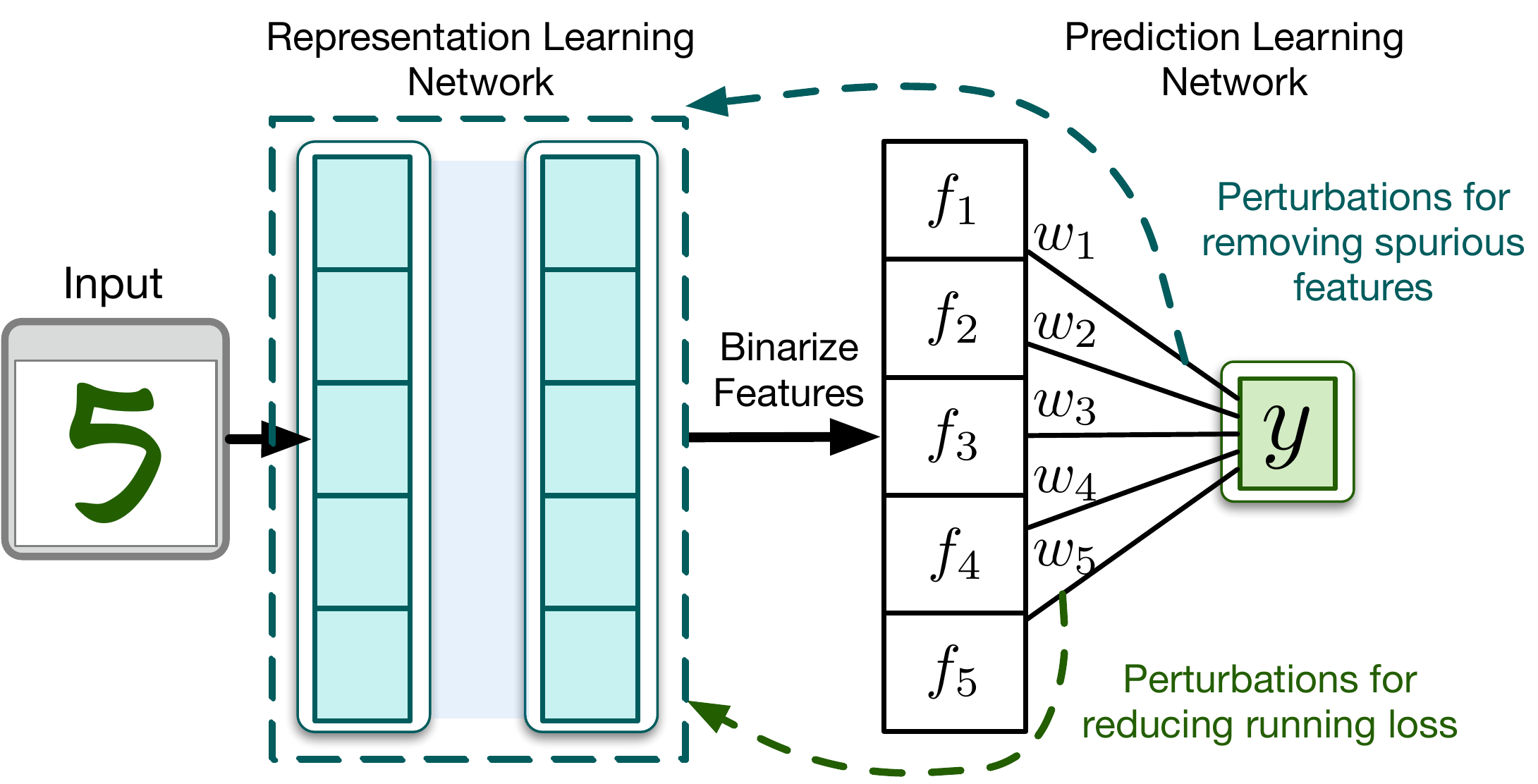}
	\end{center}
	\caption{The overall architecture of our learner. The agent learns a linear predictor on binary features $f_i$ using weights $w_i$ online. The agent also maintains statistics about the weights, such as their variance and magnitude, over time. Based on these statistics, it changes the parameters in the representation learning network using direct feedback mechanisms. Changes are kept if they improve the learning metrics, and reverted back otherwise. Given statistics that capture the performance of the learner and degree of spurious correlations between the features and the target, our architecture pushes the learner to learn a set of features that are highly correlated with the target, but not in a spurious way.}
  \label{overview}
	\vspace{-10pt}
\end{wrapfigure}

\subsection{Online Learning}
Online learning is a paradigm in which an agent continually learns as it is interacting with the world. 
This is in contrast to most of the machine learning methods that involve a separate learning and testing phase. 
An online learner has several advantages over an offline learner. First, an online learning agent does not have to learn a global predictor for the complete data-distribution. Instead, it can do \emph{tracking} -- performing better at the current part of the world even if at the expense of temporally distant parts. Tracking is not only important for practical reasons -- for complex problems, the world is much bigger than our models making learning the optimal predictor impossible -- but can also achieve better performance than a global solution for both stationary \citep{sutton2007role} and non-stationary prediction problems \citep{silver2008sample}. Second, online learning can benefit from the temporal structure of the data-stream that is often missing from offline data-sets. Finally, an online learner can do interventions -- by taking actions -- to acquire the information necessary for testing hypotheses.
\subsection{Causality}
There has been a surge in interest in bringing the machinery of causality into machine learning, with several new methods proposed over the past year. \citet{bengio2019meta} argued that causal models can adapt to interventions -- changes in distributions -- quickly and proposed a meta-learning objective that optimizes for fast adaptation. Similarly, \citet{ke2019learning} proposed a meta-objective that optimizes for models that are robust to sparse and frequent interventions. \citet{arjovsky2019invariant} took a different approach and argued that causality can be defined in terms of finding features such that the expected value of target given those features is constant across environments. They proposed Invariant Risk Minimization (IRM), a learning objective for finding such features; \citet{ahuja2020invariant, krueger2020out} expanded on IRM. 

Both categories of methods -- IRM and the one proposed by \citet{bengio2019meta} -- are incompatible with online learning. IRM \citep{arjovsky2019invariant} requires sampling data from multiple environments simultaneously for computing a regularization term pertinent to its learning objective, where different environments are defined by intervening on one or more variables of the world.
Similarly, methods proposed by \citet{bengio2019meta, ke2019learning} require sampling data before and after an intervention for computing the loss for their proposed meta-objectives. These methods can be implemented online when interventional data is temporally close -- such as the agent causing the intervention using its actions; however, oftentimes the interventional data is separated by days or even months. For example, seasonal changes provide useful interventional data for learning but happen at the span of months. Sampling simultaneously from such temporally distant parts of the world is not feasible for a practical online learner.

Nonetheless, the idea that the expected value of targets for causal features is constant \citep{arjovsky2019invariant} is interesting. We extend this idea to devise an online learning algorithm for detecting spurious features, even when the data necessary to detect the change is temporally distant. Moreover, our method does not require explicit knowledge of the type and time of intervention. This is important because a large number of interventions are unobserved and are caused by other agents, or by factors outside the control of our agent -- such as changes in weather.

\section{Problem formulation}

We look at the problem of learning to make predictions in a Markov decision process (MDP) defined by $(\statespace, \actionspace, r, p)$, where $\statespace$ is the set of states, $\actionspace$ is the set of actions, $r: \statespace \times \actionspace \times \statespace \to \real $ is a reward function, and $p:(s_t, a_t, s_{t+1}) = P(S_{t+1}=s_{t+1} | A_t=a_t, S_t = s_t)$  is the underlying transition model of the world from $ \statespace \times \actionspace \to \statespace$.  At time-step t, the agent takes an action $a_t \in \actionspace$ and the world transitions from $s_{t}$ to $s_{t+1} \in \statespace$, emitting reward $r_t$. Instead of seeing $s_{t}$ directly, the agent sees an observation $\observation_t = e(s_{t})$, an encoding of the state with an unknown encoder $e:\statespace \to \observationspace$. Encoder $e$ could be invertible -- making the observation Markovian -- or non-invertible -- requiring a recurrent mechanism for constructing agent state.
The agent state $s'_t$ -- not necessarily the same as the state of the MDP  -- is the same as $\observation_t$ if $e$ is invertible. Otherwise, $s{'}_t = U(s{'}_{t-1}, \observation_{t})$, where $U$ is the state-update function. Our notation follows the standard set by \citet{sutton2018reinforcement}.

In a prediction problem, the agent has to learn a function $f_\theta(s^{'}_t, a_t)$ to predict a target $y_t$ using parameters $\theta$. As the agent transitions to the new state $s_{t+1}$, it receives the ground truth label $\hat{y}_t$ from the environment and accumulates regret given by $\mathcal L(y_t, \hat{y}_t)$, where $\mathcal L$ is a loss function that returns the prediction error, such as mean squared error. The agent can use $\hat{y}_t$ to update its estimate of $f_\theta(s^{'}_t, a_t)$. This formulation can represent important prediction problems, such as learning a model for model-based RL, online self-supervised learning, or learning General Value Functions (GVFs) \citep{sutton2017horde}. 

The goal of the agent is to learn $f_\theta(s^{'}_t, a_t)$ such that the learned function generalizes to unseen parts of the MDP. Such generalization is important because the agent might be interested in counterfactual reasoning or planning for taking actions in unseen parts of the world. For example, the agent might want to decide against jumping off a cliff without ever trying it once. To do so, the agent must learn to predict the outcome of the fall without ever attempting it by generalizing from prior experience. 

\section{An online algorithm for identifying spurious features}

Consider $n$ features $x\defeq f_1, f_2,\cdots,f_n$ that can be linearly combined using parameters $w_1, w_2,\cdots,w_n$ to predict a target $y$. Moreover, assume all features are binary -- 0 or 1.  Given these features, our goal is to identify and remove the spurious ones.
We define a feature $f_i$ to have a spurious correlation with the target $y$ if the expected value of target given $f_i$ is not constant in temporally distant parts of the MDP i.e. $\mathbb{E}[y|f_i=1]$ slowly changes as the agent interacts with the world. This is similar to the definition proposed by \citet{arjovsky2019invariant} with a key difference: Instead of introducing a notion of multiple environments, we aim to find features with stable correlation across temporally distant parts of the same MDP. We also avoid defining features with stable correlations as causal ones -- It is possible that by exploring new parts of the world, the stable features might also turn out to be spurious.

Given this definition, we first propose an online algorithm for detecting spurious features from a given set of features. For a linear prediction problem, detecting if the $ith$ feature is spurious is equivalent to tracking the stability of the $w_i$ across time. i.e. if the online learner is always learning using the most recent data with the following update:
\begin{equation}
 \theta^{t} = \theta^{t-1} - \gamma  \nabla_{\theta^{t-1}}\mathcal L(f_{\theta^{t-1}}(s^{'}_t, a_t), \hat{y}_t) ,
 \label{update_rule}
\end{equation}
the weight corresponding to the features with a constant expected value, $\mathbb{E}[y|f_i=1]=c$,  would converge to a fixed magnitude. Whereas if $\mathbb{E}[y|f_i=1]$ is changing, $w_i$ would track this change by changing its magnitude overtime. This implies that weights that are constantly changing in a stationary prediction problem encode spurious correlations. We can approximate the change in the weight, $w_i$, overtime by approximating its variance online. Our hypothesis is that spurious features would have weights that have high variance.

To approximate variance online, we keep two exponentially decayed sums for each feature. First, we keep track of the running mean $u_i$ of the weight $w_i$ as the agent learns in the environment using Equation~\ref{update_rule}. We only update $u_i$ when $f^t_i = 1$. This is important because we only care about our estimate when a feature is active. The second metric, $v_i$, accumulates the variance of $w_i$ around the running mean $u_i$. Again, we only update $v_i$ when $f_i=1$. The update rule of both statistics is given by:

\begin{equation}
u^t_i = \alpha u^{t-1}_{i} + (1-\alpha)w^t_i f^t_i   +   (1-\alpha)(1-f^t_i)u^{t-1}_{i}
\label{mean}
\end{equation}
\begin{equation}
v^t_i = \beta v^{t-1}_{i} + (1-\beta)(w^t_i - u^t_i)(w^t_i - u^{t-1}_{i}) f^t_i   +   (1-\beta)(1-f^t_i) v^{t-1}_{i}
\label{variance}
\end{equation}
where $0 < \alpha, \beta < 1$ and $\beta < \alpha$. Equation~\ref{variance} is the Welford's method \citep{welford1962note} for computing variance online, modified to compute an exponentially decayed estimate.
\subsection{Evaluation}
\begin{wrapfigure}{R}{0.50\textwidth}

	\vspace{-40pt}
	\begin{center}
		\includegraphics[width=0.50\textwidth]{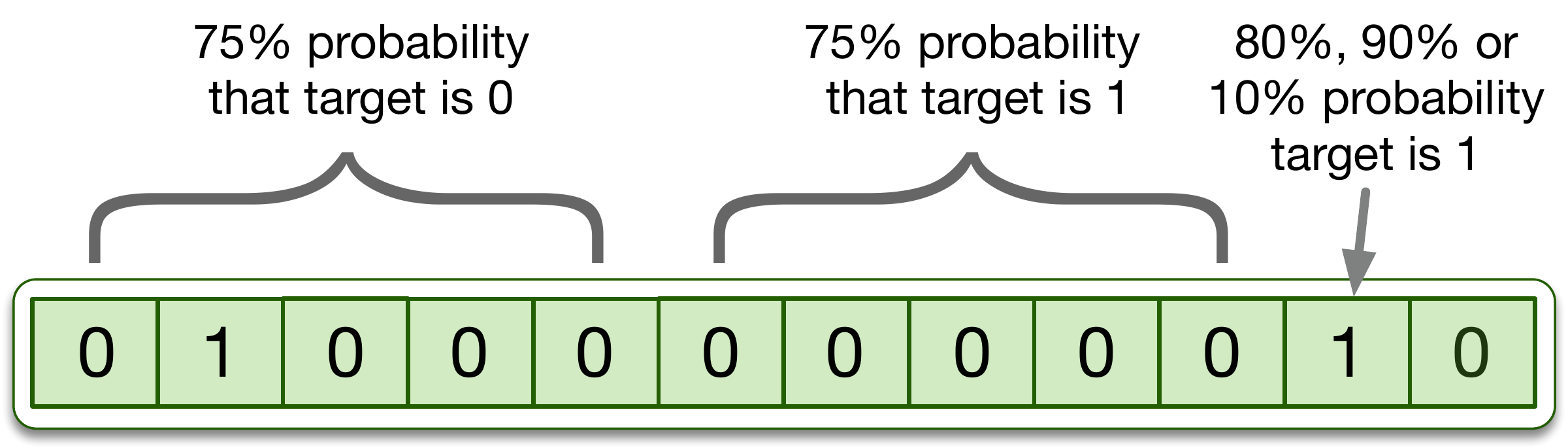}
	\end{center}
	\caption{Feature space of the Online Colored-MNIST benchmark. The last two features -- encoding color -- strongly correlate with the target y in parts of the environment used for learning. However, the degree of correlation changes over time. During learning, the agent only explores the part of the state space where color can predict the target with 80 or 90 percent probability. We evaluate this predictor on the part of the MDP where this correlation is reversed. Even though this problem does not require feature learning, ignoring the highly correlated color label in an online setting is not trivial.}
	\label{feature}
	\vspace{-29pt}
\end{wrapfigure}

We first verify that the variance metric can detect features that are spurious with high confidence in a simple setting. We then extend the algorithm with a representation search to also learn causal features online. 

\subsubsection{Benchmark}
We design an online binary classification benchmark -- Online Colored MNIST -- formulated as an MDP; The first five classes of MNIST -- 0,1,2,3,4 -- correspond to target 0 whereas the remaining five correspond to target 1. We flip 25\% of the targets at random to introduce noise. Every digit is written in green or red ink as shown in Figure~\ref{overview}. The color of the digit strongly correlates with the target in a spurious way in some parts of the MDP i.e. $\mathbb{E}[y=1|\textit{color}=\textit{green}]$ is 0.8, 0.9, or 0.1 depending on the state of the MDP. At every state, the agent receives an observation -- a set of features describing the partial state of the MDP. The observation consists of a one-hot encoded class label and one-hot encoded background color as shown in Fig.~\ref{feature}. The agent can take only one action that changes the label of the image from $x$ to $x+1$, $x+2$, $x+3$, $x+4$, or $x+5$ (Modulo $10$) with $15\%, 10\%, 5\%, 3\%,$ and $2\%$ probability, respectively. The class label remains unchanged with $65\%$ probability. Moreover, with $0.01\%$ probability,  $\mathbb{E}[y=1|\textit{color}=\textit{green}]$ changes from 0.9 to 0.8 and vice-versa. The agent is evaluated on a different part of the MDP where $\mathbb{E}[y=1|\textit{color}=\textit{green}]$ is 0.1. The expected value of the target given color is a latent variable not observed by the agent. We test if our algorithm can discover that the correlation between color and target is spurious by only learning on parts of the MDP where $\mathbb{E}[y=1|\textit{color}=\textit{green}]=0.8$ or $0.9$. To evaluate the algorithm, we freeze learning and drop the agent on the part of the MDP  where $\mathbb{E}[y=1|\textit{color}=\textit{green}]=0.1$. We call the part of the MDP used for learning \textit{Seen MDP} and the one used for evaluation \textit{Unseen MDP} To perform well, the agent has to generalize to the \textit{Unseen MDP} in a zero-shot way. An agent that relies on the spurious correlation -- the background color -- for making predictions would generalize poorly. Our benchmark is an online version of the Colored MNIST benchmark proposed by  \citet{arjovsky2019invariant}.

\subsubsection{Baselines} 
We compare our method with the following online and offline learning baselines. All methods learn a linear model from features to target.
\begin{wrapfigure}{R}{0.60\textwidth}

	\vspace{-9pt}
	\begin{center}
		\includegraphics[width=0.60\textwidth]{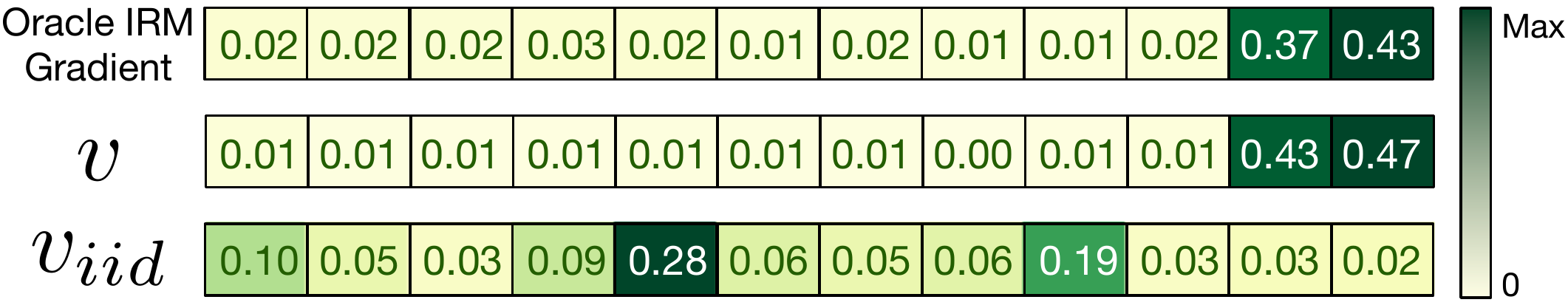}
	\end{center}
	\caption{Comparing online metric $v$ and the gradient of the penalty term used by the Oracle IRM. Large value of $v$ and the gradient indicates a spurious feature. Both $v$ and Oracle IRM can identify the last two features -- encoding color -- as spurious. However, unlike $v$, Oracle IRM requires samples conditioned on the latent variable to work. We also note that $v_{iid}$ fails to detect spurious features implying that the temporal structure in the data is essential for detecting spurious features.}
	\label{v_vs_gd}
	\vspace{-9pt}
\end{wrapfigure}

\paragraph{Online Learning}
The learner uses Equation~\ref{update_rule} for learning by minimizing risk on the most recent sample.
\vspace{-10pt}
\paragraph{Oracle IRM} 
The agent uses the IRM objective \citep{arjovsky2019invariant} for learning. We fix the weights $w_i$ to 1, as done in the IRM paper, and instead learn gating weights $g_i$ for features $f_i$. IRM learns $g_i$ using a sum of two different gradients. First, it computes loss, $L_1$, using Equation~\ref{update_rule} for a large batch of data sampled IID from the \textit{Seen MDP}. Second, it computes the gradient of the loss with two batches -- one for which $\mathbb{E}[y=1|\textit{color}=\textit{green}] = 0.8$ and another for which it is 0.9. It then squares these gradients individually, sums them, and minimizes this sum in addition to minimizing $L_1$ (weighted by a hyper-parameter). Since IRM requires samples conditioned on the hidden variable -- the correlation of target with the color -- for computing the penalty term, we call it Oracle IRM. The other methods do not use this information. 
\vspace{-10pt}
\paragraph{Online IRM}
We use the same objective as IRM, but compute the gradient penalty term explained above using only the most recent sample.
\vspace{-10pt}
\paragraph{Our method}
We compute $v_i$ and $u_i$ for every feature $f_i$ as the agent is interacting with the world using Equation~\ref{mean}~and~\ref{variance}. We use $\alpha=0.999$ and $\beta=0.9999$. Moreover, we initialize a new parameter, $g_i$, and initialize it to be zero. We mask the feature $f_i$ using $\sigma(g_i)$, where $\sigma$ is the sigmoid function. After sufficient learning -- 500,000 steps --  we update $g_i$ as $g_i = g_i - \frac{1}{|v|^2_2}(v_i  - \frac{1}{n}\sum_{j=1}^{n} v_j)$ every 50,000 steps. We could also use $g = g - softmax(v)$ for the update here. Our update decreases $g_i$ if $v_i$ is above the average value of $v$, and increases it otherwise. Since we expect spurious features to have a higher than average variance, our update rule should push $\sigma(g_i)$ to zero if $f_i$ is spurious.

\begin{wraptable}{R}{0.50\textwidth}
	\vspace{-20pt}
	\centering
	\caption{Classification accuracy on the seen and unseen part of the MDP. All methods first learn the prediction function $f$ using one million steps. Learning is then stopped and the agents are evaluated on the seen and unseen parts of the MDP for 100,000 steps. Our method is the only one that can learn to ignore spurious correlations online and matches the performance of Oracle IRM on unseen parts of the MDP. All numbers represent percentage classification accuracy over 100,000 steps and the error margins represent one std computed using bootstrapping.}
	\begin{minipage}{\textwidth}
		
			\begin{tabular}{ccc}
				\toprule
				\centering
				Method &   Seen MDP & Unseen MDP  \\
				\midrule
				Online Learning & 85.03 {\tiny $\pm$  0.03} & 10.01 {\tiny $\pm$  0.01}\\ 
				Oracle IRM & 75.01 {\tiny $\pm$  0.02}  & 75.00 {\tiny $\pm$  0.02}  \\ 
				Online IRM & 85.01 {\tiny $\pm$  0.02}  & 09.99 {\tiny $\pm$  0.02} \\ 
				Our Method & 75.00 {\tiny $\pm$  0.02}  & 75.01 {\tiny $\pm$  0.02}  \\ 
				\bottomrule
			\end{tabular}
	\end{minipage}
	\label{features}
	\vspace{-20pt}

\end{wraptable}

For each method, we use the Adam optimizer \citep{kingma2014adam} for online learning for one million steps using logistic regression to predict the target; Moreover, we do a grid search over the hyper-parameters -- over the learning rate and regularization strength for the parameters -- and report the best results for each method. For Oracle IRM and our method, we tried learning rates in set $(10^{-3}, 10^{-4},10^{-5})$ and $l_1$ regularization in set $(10^{-2}, 10^{-3}, 10^{-4})$. For Online IRM and online learning baseline, we tried both $l_1$ and $l_2$ regularization and did a bigger sweep: from $10^2$ to $10^{-7}$ for both the learning rate and the regularization term. Pseudo-code for Oracle IRM and our method, additional implementation details, and link to the executable code is in the appendix. 
\subsubsection{Results}

To verify that our method can identify spurious correlations, we first compare the $v$ estimate with the gradients computed using the regularization penalty of Oracle IRM\footnote{We compute the IRM regularization term gradients for network pre-trained on the complete distribution till convergence to remove noise.}. We also add another baseline $v_{iid}$ computed exactly as $v$ except the data is first stored in an experience replay buffer of size 500,000 and then sampled IID for learning. We normalize all three so they sum to one and plot them in Figure~\ref{v_vs_gd}. $v$ captures the same information as Oracle IRM but in an online way without using information from the latent variable. Both the Oracle IRM gradients and $v$ can detect spurious features with high confidence. Moreover, shuffling data -- by sampling IID from a buffer -- destroys the necessary information for detecting spurious features online.

We then let all the methods learn for one million steps in \textit{Seen MDP} and report the results for both \textit{Seen MDP} and \textit{Unseen MDP} in Table~\ref{feature}. For Oracle IRM, we learn by sampling IID until convergence. Both oracle IRM and our method achieve strong zero-shot generalization by learning to ignore the spurious features robustly across hyper-parameter settings. On the other hand, online learning and online IRM fail to identify spurious correlations regardless of the large hyper-parameters search. For every run, these methods either (a) did not learn anything getting $\sim50\%$ accuracy for both Seen and Unseen MDP when the regularization term was too strong, or (b) converged to a solution that relied on the color information for making predictions.
\vspace{-10pt}
\paragraph{Can experience replay help online IRM?}
Experience replay buffers fix some issues with online learning by storing past $K$ examples and sampling IID from these examples for learning \citep{lin1993reinforcement, mnih2015human}. We combine online IRM with a sufficiently large buffer and report the results in Table~\ref{buffer}. The IRM objective, even when combined with an experience replay buffer, is still unable to detect spurious features. This highlights that the poor performance of online IRM is not due to the instability of online learning; IRM needs samples conditioned on the latent variable --  $\mathbb{E}[y=1|\textit{color}=\textit{green}]$ -- to work.

\begin{wraptable}{R}{0.45\textwidth}
	\vspace{-60pt}
	\centering
	\caption{Even with an experience replay, the IRM penalty is unable to detect spurious features. This is perhaps not surprising, as sampling IID from an experience replay buffer throws away the temporal information in data necessary for detecting spurious features. All numbers represent percentage classification accuracy.}
	\vspace{3pt}
	\centering
	\begin{minipage}{\textwidth}
		
			\begin{tabular}{ccc}
				\toprule
				\centering
				Buffer size &   Seen MDP & Unseen MDP  \\
				\midrule
				100,000 & 85.01 {\tiny $\pm$  0.02} & 10.00 {\tiny $\pm$  0.02}\\ 
				500,000 & 85.02 {\tiny $\pm$  0.02} & 09.98 {\tiny $\pm$  0.02}\\ 
				\bottomrule
			\end{tabular}
	\end{minipage}
	\label{buffer}
	\vspace{-8pt}

\end{wraptable}

\section{Online Feature Discovery}
The previous experiments provide evidence that the online variance metric identifies spurious features. In this section, we show that we can use $v$ to also learn features for sensory data. Because Equation~\ref{mean} and \ref{variance} are differentiable, we could use gradient-based learning to compute gradient through the update equation for $v$ -- similar to how gradient-based meta-learning methods compute gradients through SGD updates \citep{santoro2016meta, finn2017model, li2017meta}. However, getting an accurate estimate of $v$ can take thousands to millions of steps, depending on the mixing time of the data-stream; this makes computing the true gradient using BPTT \citep{werbos1990backpropagation} intractable. Gradient-based learners explicitly designed for long-distant credit assignment \citep{ke2018sparse} also require storing network activations from previous data-points, making them impractical for our setting. Some work has proposed approximating the true gradient by only computing gradients through past $K$ steps \citep{williams1990efficient, sutskever2013training, javed2019meta}; however, these approximations are incapable of effectively capturing a spurious correlation if the data necessary for detecting this correlation is more than $K$ steps apart. 
\subsection{Perturbations with backtracking}
To avoid the issues associated with gradient-based learning for long-distant credit assignment, we propose to do a weakly directed search in the parameter space for learning features. We divide our network into two parts -- a Representation learning network (RLN) and a Prediction learning network (PLN) as shown in Figure~\ref{overview}. Our learner learns the PLN online using Equation~\ref{update_rule}. It also maintains an exponentially decayed estimate of loss -- regret -- and the $v$ estimate for the weights in the PLN. 

For learning, the learner perturbs some weights of the RLN by setting them to $0$, $+1$, or $-1$. After a perturbation, the learner observes the running loss and $v$ as it continues to update PLN. A decrease in the running loss indicates that the feature after the perturbation is a better predictor of the target $y$. Similarly, if the sum of $v$ is reduced after the perturbation, the new features are less spuriously correlated with the target than before. If either the running loss or sum of $v$ is decreased after the perturbation, the perturbation is kept. Otherwise, the agent reverts to the older parameters. Because a perturbation is only kept if it improves one of the metrics -- by reducing loss or by reducing $v$ -- the learner is guaranteed to improve or retain its performance. We call this method Perturbations with Backtracking (PwB).

Random search has been explored as a learning mechanism in the past. \citet{li2019random} and \citet{mania2018simple} found random search to be a strong baseline for neural architecture search and linear control, respectively. \citet{mahmood2013representation, mahmood2017incremental} proposed \textit{Generate and Test} as a mechanism for learning representations. Their method measured the importance of each feature online and replaced the least important features with new random features. Our method differs from theirs by using a mechanism for backtracking if a perturbation is detrimental. We found backtracking to be crucial for making consistent improvements.

\begin{wraptable}{R}{0.50\textwidth}
	\vspace{-6pt}
	\centering
	\caption{Percentage accuracy for image-based Online Colored MNIST averaged over 10 runs. We report the results of two different versions of IRM. IRMv1 achieves the best result while not relying on any spurious features, whereas v2 achieves the highest average accuracy. Both Oracle IRM and PwB* can learn to ignore spurious correlations from raw data, but only PwB* can be implemented online. Surprisingly, PwB*, performs better than Oracle IRM; however, we suspect it would be possible to carefully tune Oracle IRM to get similar results as well.}
	\begin{minipage}{\textwidth}
		
			\begin{tabular}{ccc}
				\toprule
				\centering
				Method &   Seen MDP & Unseen MDP  \\
				\midrule
				Oracle IRMv1 & 64.78 {\tiny $\pm$  0.08}  & 63.87 {\tiny $\pm$  0.07}   \\ 
				Oracle IRMv2 & 73.76 {\tiny $\pm$  0.07}  & 63.10 {\tiny $\pm$  0.01}   \\ 
				PwB* & 68.83 {\tiny $\pm$  0.49}  & 68.73 {\tiny $\pm$  0.67}   \\ 
				PwB* (0.85) & 84.01 {\tiny $\pm$  1.00}  & 13.99 {\tiny $\pm$  5.57}   \\ 
				\bottomrule
			\end{tabular}
	\end{minipage}
	\label{images}
	\vspace{-44pt}

\end{wraptable}

\subsection{Benchmark}
We use the same Online Colored MNIST benchmark but learn directly from images instead of binary features. We make the difference in $\mathbb{E}[y=1|\textit{color}=\textit{green}]$ more extreme by varying the value from 0.76 to 0.99 instead of 0.8 to 0.9. A higher difference allows both Oracle IRM and PwB to remove the spurious correlation more robustly across hyper-parameters. For real-life spurious correlations, we expect these differences to be even larger. 
\subsection{Implementation Details and Results}
 We use a one convolution layer followed by a fully connected layer to get a set of features. For PwB, we binarize these features by treating positive values as one, and non-positive values as zero. For Oracle IRM, we use ReLU non-linearity instead so that it is differentiable.
 
 We take turns minimizing running loss and minimizing $v$ for PwB. For minimizing $v$, we perturb the parameters in the convolution layers -- by setting 0.001 to 0.3 percent of parameters to 0 or +1 and seeing if the perturbation reduces the sum of $v$. If it does, the perturbation is kept. For minimizing the loss, we perturb the weights in the fully connected layer by setting 0.001\% to 0.3\% of parameters to 0, +1, or -1 and keeping the changes if the running loss decreases over-time. By perturbing different layers for the two metrics, we avoid competition between them. An alternative would have been to perturb all layers simultaneously to minimize a weighted sum of $v$ and the loss. 
 
 PwB requires evaluating a perturbation by relearning the last layer predictor until convergence. This can be done online but can be slow for running experiments. To speed up the experiments, we cheat by computing the new value of $v$ after a perturbation offline by sampling batches of data conditioned on  $\mathbb{E}[y=1|\textit{color}=\textit{green}]$, similar to Oracle IRM. We fit two different linear predictors on the two batches, one for which the latent variable is 0.76 and the other for which it is 0.99. We subtract one set of resulting weights from the other, square the resulting vector, and sum it. This gives us an offline estimate of $v$ that can be computed faster. Similarly, we use an offline estimate of the running loss by sampling IID from the Seen MDP. To confirm that the online metric is also capable of capturing the same information, we compute the Pearson's correlation coefficient between the sum of $v$ for the online and offline estimate for 100 different perturbations and found it to be +0.91; the strong correlation indicates that the online estimate should give similar results. We label PwB that uses the offline estimate of $v$ as PwB*.

 We report the results of PwB* and Oracle IRM in Table~\ref{images}. We did a large grid search for Oracle IRM, trying 50 different configurations, and report results for the two best configurations. Both Oracle IRM and PwB* can learn to remove spurious correlations; however, they have a key difference -- PwB* can be implemented online whereas Oracle IRM is inherently incompatible with online learning. As a sanity checks, we fix the latent variable to 0.85 for all states in the Seen MDP and run PwB* using the same hyper-parameters. We call this variant PwB* (0.85). Because the correlation of color with the target is stable now, PwB* (0.85) should use the color information for making predictions. We confirm that this is indeed the case in Table~\ref{images}.  For more implementation details, pseudo-code for PwB, and link to the executable code, see the appendix.  
 \section{Conclusion}
 In this work, we proposed an online estimate that can be used to identify spurious features; unlike earlier methods for causal learning, our method is scalable and does not require information about the source and time of interventions; it can also learn non-spurious features from raw sensory data. Moreover, our representation search method avoids the limitations of gradient-based learning and is capable of credit assignment across long durations of time.
 
Our work also has one key limitation -- it uses random search for creating perturbations which can be sample-inefficient in large parameter spaces. There are several ways the search can be made more efficient; we could use networks that are sparsely connected, reducing the number of parameters; alternatively, instead of searching for the parameters directly, we could search for direct feedback paths for targeted random perturbation. We could also bias the search using heuristics. Finally, we could equip our online learner with a curriculum to guide the search to solve complex problems by solving easier problems first. All of these are interesting venues for furthering the ideas presented in this paper.

\bibliographystyle{chicagoo}
\bibliography{interleaved}

\newpage
\appendix
We provide implementation details of the experiments in Section 3 and 4 in the following sections. A copy of the executable code is also available \footnote{\url{https://github.com/khurramjaved96/online-causal-models}}.

\section{Feature selection experiments} 

\begin{figure}[h]
	
		\begin{algorithm}[H]
			
			\centering
			\begin{algorithmic}[1]
				\caption{Feature Selection: Oracle IRM}
			\REQUIRE  Distribution over inputs $\mathcal X$ and targets $\mathcal Y$;
			\REQUIRE $s$: Total learning steps. $f_\theta$: function to learn;
			\REQUIRE $w$: Warm up steps, $\mathcal L$: Loss function for the prediction error;
			\REQUIRE $\gamma$: Learning rate; $r$: regularization weight; $p$: IRM penalty; weight.
			\REQUIRE Features: $x= (f_1, f_2, \cdots f_n)$; gating weights $\theta = (g_1, g_2, \cdots, g_n)$;
			\REQUIRE Classifier weights $(w_1, w_2, \cdots w_n)$;
			\STATE Initialize $w_i=1$ and $g_i=1$ for $i$ from $1$ to $n$.
				\FOR{$i = 1, 2, \cdots, s$}
				\STATE Sample batch $\vec{x_{0.8}}, \vec{y_{0.8}}$ \hfill  \#Sampling conditioned on $\mathbb{E}[y=1|\textit{color}=\textit{green}]=0.8$  
				\STATE Sample batch $\vec{x_{0.9}}, \vec{y_{0.9}}$ \hfill \#Sampling conditioned on $\mathbb{E}[y=1|\textit{color}=\textit{green}]=0.9$
				\STATE Sample batch $\vec{x}, \vec{y}$ \hfill \#Sampling uniformly from the MDP 
				\STATE $l_{irm}^{0.8}$ = ComputeIRMPenalty($f_\theta(\vec{x_{0.8}}), \vec{y_{0.8}} $)  \hfill \#IRM loss on conditioned data.
				\vspace{3pt}
				\STATE $l_{irm}^{0.9}$ = ComputeIRMPenalty($f_\theta(\vec{x_{0.9}}), \vec{y_{0.9}} $)	\hfill \#See IRM paper for details
				\STATE $l_1 = ||\vec{\theta} ||_{1}$ \hfill \#$l_1$ penalty loss
				\STATE $l_{pred} = \mathcal L(f(\vec{x}), \vec{y})$ \hfill \#Predictor error.
				\IF{$i > w$}
				\STATE $l_{final} = l_{pred} + rl_1 + p(l_{irm}^{0.8} + l_{irm}^{0.9})$ \hfill \#Combined loss
				\ELSE
				\STATE  $l_{final} = l_{pred} + rl_1$ \hfill \#Only apply IRM penalty loss after $w$ warm-up learning steps. 
				\ENDIF
				
				\STATE $\theta =\theta - \gamma \nabla\theta l_{final}$ 
				\ENDFOR
			\end{algorithmic}
			\label{oracle_irm}
		\end{algorithm}
		
\end{figure}

\begin{wraptable}{R}{0.70\textwidth}
	\vspace{-20pt}
	\centering
	\caption{Hyper-parameters tried for feature selection by our method and Oracle IRM}
	\begin{minipage}{\textwidth}
		
			\begin{tabular}{ccc}
				\toprule
				\centering
				Method &   Oracle IRM & Our method  \\
				\midrule
				Learning rate ($\gamma$) &  $10^{-3}, 10^{-4}, 10^{-5}$  & $10^{-3}, 10^{-4}, 10^{-5}$   \\ 
				$l_1$ strength ($r$) & $10^{-2}, 10^{-3}, 10^{-4}$  & $10^{-2}, 10^{-3}, 10^{-4}$    \\ 
				Mask lr ($p$) & N/A &  $10^{-3}, 10^{-4}, 10^{-5}, 10^{-6}$   \\ 
				IRM penalty ($p$) & $10^3, 10^4, 10^5, 10^6$  & N/A   \\ 
				\bottomrule
			\end{tabular}
	\end{minipage}
	\label{params}
	\vspace{-7pt}

\end{wraptable}

Pseudocode for feature selection algorithms -- Oracle IRM and our method -- is given in Algorithm~\ref{oracle_irm} and \ref{our_method}, respectively. We used $s = 5,000,000$ for our method and $s=1,000$ for Oracle IRM. Since Oracle IRM uses a mini-batch for 1024 for every update whereas our method uses a single sample, both methods use a comparable number of examples for learning (Oracle IRM uses a bit more, in fact). $w$ equals 2,000 and 3,000,000 for Oracle IRM and our method, respectively. Both methods use the binary cross-entropy loss for learning and take less than two hours on a single CPU to converge to the optimal solution. 

For both methods, we did a hyper-parameter sweep over the remaining important parameters -- learning rate and regularization strength. For Oracle IRM, we also did a sweep over IRM penalty weight whereas, for our method, we did a sweep over the mask learning rate (Line 12 in Algorithm~\ref{our_method}). We tried 36 different combinations of parameters for both methods as described in Table~\ref{params}. Both methods robustly converged to the optimal solution for most of these configurations (Oracle IRM failed for 8 of them whereas our method for only 4).  

For Online IRM and Online Learning, we did an even larger sweep over parameters. The two methods did not learn to ignore the spurious features in any of the runs. 

\begin{figure}
	
		\begin{algorithm}[H]
			
			\centering
			\begin{algorithmic}[1]
				\caption{Online Feature Selection: Our Method}
			\REQUIRE  m: MDP that takes action $a_t$ as input and returns $x_{t+1}$ and $\hat{y_t}$;
			\REQUIRE $s$: Total learning steps; $f_\theta$: function to learn;
			\REQUIRE $w$: Warm up steps; $\mathcal L$: Loss function for the predictor error;
			\REQUIRE $\gamma$: Learning rate; $r$: regularization weight; $p$: Mask update weight;
			\REQUIRE $\alpha$: Mean decay rate; $\beta$: Variance decay rate;
			\REQUIRE $x_1$: Initial agent state; $s_i$: Initial MDP state $r$: regularization weight; $p$: mask learning rate;
			\REQUIRE Features: $\vec{x}= (f_1, f_2, \cdots f_n)$; mask weights $ \vec{M} = (m_1, m_2, \cdots, m_n)$;
			\REQUIRE Classifier weights $\vec{\theta} = (w_1, w_2, \cdots w_n)$;
			\STATE Initialize $w_i=0$ and $m_i=0$ for $i$ from $1$ to $n$.
			\STATE Initialize mean $\vec{u} = (u_1, u_2, \cdots u_n) = \vec{0}$
			\STATE Initialize variance $\vec{v} = (v_1, v_2, \cdots v_n) = \vec{0}$
			\FOR{$i = 1, 2, \cdots, s$}
			\STATE $\vec{x_{i+1}}, \hat{y_i} = m(a)$ \hfill \# Take action $a$ and get observation and previous target from the MDP
			\STATE $\theta = \theta - \gamma\nabla \theta \mathcal L(f(\vec{x_i}\sigma(\vec{M})), \hat{y}_i))$ \hfill \# Update linear predictor online. Mask features using $M.$
			\STATE $\vec{\theta} = \vec{\theta} - r\nabla \theta ||\vec{\theta}||_1$ \hfill \# $l_1$ regularization.
			\STATE $\vec{u_{old}} = \vec{u}$
			\STATE $\vec{u} = \alpha \vec{u} + (1-\alpha)\vec{\theta} \vec{x_i}   +   (1-\alpha)(1-\vec{x_i})\vec{u}$ \hfill \# Equation 2 for online mean estimate
			\STATE $\vec{v} = \beta \vec{v} + (1-\beta)(\vec{\theta} - \vec{u})(\vec{\theta} - \vec{u_{old}}) x_i   +   (1-\beta)(1-\vec{x_i}) \vec{v}$ \hfill \# Eq 3 for variance estimate
			\IF{$i > w$}
			\STATE $\vec{M} = \vec{ M} - p\vec{v}$ \hfill \#Updating mask for hiding spurious features
			\ENDIF
			\ENDFOR
			\end{algorithmic}
			\label{our_method}
		\end{algorithm}
\end{figure}

\section{Feature discovery experiments} 
The PwB algorithm is described in Algorithm~\ref{pwb}.
MNIST images are down-sampled to 14x14 for all image-based experiments, similar to the original IRM paper \citet{ahuja2020invariant}. Pseudocode for PwB is given in Algorithm~\ref{pwb_algo}. The fitPLN function in Algorithm~\ref{pwb_algo} can be implemented online or offline. PwB* implements it offline, using Ridge Regression to find the optimial linear predictor on two large batches of data sampled after conditioning on the latent variable. The online version should give similar results, but would take much longer to run. We found a strong correlation of 0.91 between the online and offline running loss and variance. 

\subsection{Network Architecture}
We used two hidden layer neural networks. The first layer consists of four $3\times3$ convolution filters applied with a stride of 1. The input image is padded by zeros on each side -- turning the 14x14 image to 16x16 -- before applying the filter. The result of the filter -- a 784 dimension vector -- is passed to a fully connected layer of dimension 100. The resulting 100-dimensional feature vector is used for learning the linear function. For PwB, the feature vector is binarized -- positive values are changed to 1 and non-positive to 0. For Oracle IRM, on the other hand, we used relu activation as binarization is not differentiable.

\begin{figure}[h]
	
		\begin{algorithm}[H]
			
			\centering
			\begin{algorithmic}[1]
				\caption{PwB}
			\REQUIRE  m: MDP that takes action $a_t$ as input and returns $x_{t+1}$ and $\hat{y_t}$;
			\REQUIRE $s$: Total learning steps; $f_W$: Prediction learning network;
			\REQUIRE $\phi_{\theta}$: Representation Learning Network;
			\REQUIRE $\gamma$: Learning rate; $r$: regularization weight;
			\REQUIRE $\alpha$: Mean decay rate; $\beta$: Variance decay rate;
			\REQUIRE $r$: regularization weight;
			\STATE Initialize PLN parameters: $W=\vec{0}$;
			\STATE Initialize RLN parameters: $\theta$ randomly;
			\FOR{$i = 1, 2, \cdots, s$}
			\STATE $v_{before}, l_{before}$ = fitPLN($W, \theta, m, \gamma, \alpha, \beta, r$); \hfill \#Fitting a linear function and returning running loss and  variance
			\vspace{5pt}
			\STATE $\theta^{'} = $ perturbRLN($\theta$); \hfill \#Random perturbation changing 0.3 to 0.001 \% of parameters of $\theta$.
			\STATE $v_{after}, l_{after}$ = fitPLN($W, \theta^{'}, m, \gamma, \alpha, \beta, r$); \hfill \#Getting loss and variance after perturbation.
			\vspace{3pt}
			\IF{$i$ modulo 2 == 0 } 
			\IF{sum($v_{after}$) < sum($v_{before}$)}
			\STATE $\theta = \theta^{'}$ \hfill \#Smaller sum of $v$ indicates features are less spurious than before.
			\ENDIF
			\ELSE
			\IF{$l_{after} < l_{before}$}
			\STATE $\theta = \theta^{'}$  \hfill \#Lower loss after a perturbation indicates new features are more predictive of $y$.
			\ENDIF
			\ENDIF
			\ENDFOR
			\end{algorithmic}
			\label{pwb_algo}
		\end{algorithm}
\end{figure}

Each weight in the convolution layer for PwB* is initialized to be either 0 or 1 with equal probability whereas each weight in the fully connected layer for PwB* is initialized to be 0, +1, or -1 with equal probability. Oracle IRM, on the other hand, uses the xavier uniform initialization with a gain of 1 for all parameters. 

We also tried a two layers fully connected architecture for Oracle IRM for feature learning; the performance of the convolution architecture and fully connected architecture was comparable. 
\subsection{Hyper-parameters}
The hyperparemeters tried by Oracle IRM and PwB* are in Table~\ref{irm_image} and \ref{pwb} respectively. We observed that PwB* is more robust to hyper-parameter changes than Oracle IRM. PwB* does not depend on sensitive parameters, such as learning rate, for learning the complete network. The only iterative learning problem PwB* has to solve is the linear prediction problem on binary features for which robust solvers exist. 

\begin{wraptable}{R}{0.50\textwidth}

	\vspace{-50pt}
	\begin{minipage}{0.45\textwidth}

			\caption{Hyper-parameters search for learning features from image by Oracle IRM. The implementation details of the image based Oracle IRM can be found in \citet{ahuja2020invariant}.}
		
			\begin{tabular}{ccc}
				\toprule
				\centering
				Method &   Oracle IRM  \\
				\midrule
				$l_2$ regularization &  $10^{-2}, 10^{-3}, 10^{-4}$  \\ 
	
				IRM penalty  & $10^3, 10^4, 10^5, 10^6$    \\
				Penalty annealling & 10, 100, 1000 steps \\ 
				Architecture 1& Two FC Layer \\
				Architecture 2& Conv + FC Layer \\
				Feature dim & 50, 100, 200 \\
				\bottomrule
				\label{irm_image}
			\end{tabular}
	\end{minipage}

	\begin{minipage}{0.45\textwidth}

			\caption{Hyper-parameters tried for learning features from images by PwB*.}
		
			\begin{tabular}{ccc}
				\toprule
				\centering
				Method &   PwB* \\
				\midrule
				$l_2$ regularization &  $10^{-2}, 10^{-3}, 10^{-4}$  \\ 
				Feature dim & 50, 100, 200 \\
				\bottomrule
				\label{pwb}
			\end{tabular}
	\end{minipage}
	

\end{wraptable}

\section{Compute resources}
Feature selection, and PwB* experiments were done on 48 core CPU servers whereas image based Oracle IRM experiments were done using V100 GPUs. The feature selection experiments can run on a single CPU core in less than two hours. For PwB*, we regress to the targets using sklearn's Ridge regression\footnote{\url{https://scikit-learn.org/stable/modules/generated/sklearn.linear_model.Ridge.html}}, which can benefit from all 48 cores of the server; using all 48 cores, PwB* converges in 3-4 hours.  

\end{document}